\documentclass[11pt]{article}

\usepackage{booktabs}
\usepackage{graphicx}
\usepackage{amsmath}
\usepackage{amssymb}
\usepackage{tikz}
\usepackage{subfigure}
\usepackage{xcolor}
\usepackage{color}
\usepackage{tcolorbox}
\usepackage{diagbox}
\usepackage{mathtools}
\usepackage{url}
\usepackage{bm}
\usepackage{arydshln} 
\usepackage{tablefootnote}
\usepackage{longtable}
\usepackage{pifont}
\usetikzlibrary{shadows}
\usetikzlibrary{shadows.blur}
\usepackage{acl}
\usepackage{times}
\usepackage{latexsym}

\usepackage[T1]{fontenc}

\usepackage[utf8]{inputenc}

\usepackage{microtype}

\title {Consecutive Question Generation via Dynamic Multitask Learning}

\author{Yunji Li\textsuperscript{1,2}, Sujian Li\textsuperscript{1}, Xing Shi\textsuperscript{2} \\
  \textsuperscript{1}MOE Key Lab of Computational Linguistics, Peking University, China \\
  \textsuperscript{2}ByteDance Lark Search \\
  \texttt{\{liyunji.0529, shixing.xingshi\}@bytedance.com} \\
  \texttt{lisujian@pku.edu.cn}}

\begin{document}
\maketitle
\begin{abstract}
In this paper, we propose the task of consecutive question generation (CQG), which generates a set of logically related question-answer pairs to understand a whole passage, with a comprehensive consideration of the aspects including accuracy, coverage, and informativeness.
To achieve this, we first examine the four key elements of CQG, i.e., question, answer, rationale\footnote{The sentence based on which a question is generated.}, and context history\footnote{The coverage of all previous rationales, representing the background information of the current question series.}, and propose a novel dynamic multitask framework with one main task generating a question-answer pair, and four auxiliary tasks generating other elements.
It directly helps the model generate good questions through both joint training and self-reranking.
At the same time, to fully explore the worth-asking information in a given passage, we make use of the reranking losses to sample the rationales and search for the best question series globally.
Finally, we measure our strategy by QA data augmentation and manual evaluation, as well as a novel application of generated question-answer pairs on DocNLI. We prove that our strategy can improve question generation significantly and benefit multiple related NLP tasks.
\end{abstract}

\section{Introduction}
Question Generation (QG) is an important and promising task in natural language generation (NLG). It has long served as an effective way to improve other NLP tasks. The applications of synthetic questions have expanded from QA data augmentation \citep{duan-etal-2017-question,patrick-etal-2021-paq} to building tutoring or dialogue systems \citep{lindberg-etal-2013-generating,antoine-etal-2017-learning}, self-assessing the ability of language models \citep{sun-etal-2019-improving}, and checking the faithfulness of an abstract summary \citep{durmus-etal-2020-feqa}, etc.

\begin{table}[t]
    \renewcommand\tabcolsep{6pt}
    \renewcommand\arraystretch{1.4}
    \centering
    \small
    \begin{tabular}{l}
    \toprule
    \textit{Today is Jessica's 80th birthday. Her daughter Mela} \\ \textit{and Mela's husband Josh is coming over to the birth-} \\ \textit{day party...} \\
    \textit{Q1:} \textcolor{cyan}{\textit{\textbf{Who is her daughter?}}}   
    \quad \quad
    \textit{A1:} \textcolor{cyan}{\textit{\textbf{Mela.}}} \\
    \textit{Q2:} \textcolor{blue}{\textit{\textbf{Who is Josh?}}} 
    \quad \quad \quad \quad \quad \ \
    \textit{A2:} \textcolor{blue}{\textit{\textbf{Mela's husband.}}}  \\
    \textit{Q3:} \textcolor{teal}{\textit{\textbf{Who has a birthday party?}}} 
    \textit{A3:} \textcolor{teal}{\textit{\textbf{Mela.}}} \\
    \bottomrule
    \end{tabular}
\caption{Example QG results using a two-step inconsecutive method based on extractive answers.}
    \label{tb1}
\vspace{-0.3cm}
\end{table}

Traditionally, syntax-based methods such as semantic parsing are commonly adopted to synthesize questions \citep{berant-etal-2013-semantic,khullar-etal-2018-automatic}. Recently, transformer-based pre-trained language models \citep{ashish-etal-2017-attention,devlin-etal-2019-bert} are widely used to generate questions.
Most of these works  are two-step QG methods \citep{sun-etal-2018-answer,rennie-etal-2020-unsupervised}, which rely on ground-truth or pre-extracted answers \citep{wang-etal-2019-multiagent,jia-etal-2020-ask} and generate questions independently \citep{puri-etal-2020-training, bartolo-etal-2021-improving}.
However, in real scenarios such as daily conversations or reading comprehension, we usually raise several questions consecutively to understand a whole story.
Current QG methods are inadequate to generate such questions, as Table \ref{tb1} shows.
We can see that there are no logical connections between the questions (e.g., Q3 and Q1) and pre-extracted answers also lead to simplicity (e.g., Q1) and inconsistency (e.g., Q3).
In such cases, we propose the task of consecutive question generation (CQG), which automatically produces a set of well-ordered and logically related question-answer (Q-A) pairs to help understand a given passage (or story). Table \ref{tb2} shows several ``ideal'' questions which are mutually connected and cover diverse information in the text. 
To achieve this, unlike traditional QG methods, which mainly focus on ``what are good questions given an answer'', our CQG also requires a model to automatically find ``which information in a text is worth-asking''. Additionally, since we pose questions not only to get separate information, but to understand a whole story, we propose three key qualities simultaneously to evaluate consecutive questions, i.e., accuracy, coverage, and informativeness.

With these demands, we propose an integrated dynamic multitask framework, with five unified Seq2Seq generation tasks. One main task generates Q-A pairs and four auxiliary tasks make full use of the generation of four key CQG elements (i.e., question, answer, rationale, and context history). We link the qualities of key aspects with the inference losses of four auxiliary tasks respectively. Based on it, we then design four distinct methods to improve the model performance from all aspects and from all stages during training and inference.

The five tasks are jointly trained in one model to help it learn from different views. In inference, the main task generates candidates and then the auxiliary tasks self-rerank them, improving Q-A accuracy, coverage, and informativeness all-roundly.
To fully exploit the worth-asking information in each sentence and generate questions properly and dynamically, we propose a novel rationale sampling method and sentence-level beam-search. We recompose the context history reranking losses to measure the information in each rationale, and then design a sample probability to guarantee that the more information a rationale leaves, the more likely it is asked once again. To relieve the error cascade and guide the direction of a Q-A flow, we reinvent beam-search to sentence-level, which rearranges the total reranking results and seeks the global optimum Q-A series for a whole passage.

\begin{table*}[!ht]
    \renewcommand\tabcolsep{6pt}
    \renewcommand\arraystretch{1.4}
    \centering
    \small
    \begin{tabular}{l}
    \toprule
    $S$: \textit{[Once upon a time in Greece, there lived a young man called Narcissus.]$^{stc_1}$ [He lived in a small} \\ \textit{village on the sea and was famous in the land because he was quite handsome.]$^{stc_2}$ ...} \\
    \midrule
    $Q_1$: What was the name of the young man? \quad \quad 
    $A_1$: Narcissus.  \qquad\qquad\qquad\quad\quad \quad \quad  $R_1$: $stc_1$ \\
    
    $Q_2$: Where did he live? \quad \quad \quad \quad \quad \quad \quad \quad \quad \quad  
    $A_2$: A small village on the sea.  \quad \quad \quad \ \ \ $R_2$: $stc_2$\\
    
    $Q_3$: Was he famous in the land? \quad \quad \quad \quad \quad \quad   \ \
    $A_3$: Yes. \qquad \qquad \qquad \quad \quad\quad\quad \quad\quad \ \ \  $R_3$: $stc_2$\\
    
    $Q_4$: Why? \quad \quad \quad \quad \quad \quad \quad \quad \quad \quad \quad \quad \quad \quad \quad 
    $A_4$: Because he was quite handsome. \quad \  $R_4$: $stc_2$\\
    \toprule
    Task \quad \quad Input \quad \quad \quad \quad \quad \quad \quad \quad \quad \quad \quad \quad \quad \quad \quad \quad \quad \quad \quad \quad \quad \quad \quad \quad \quad \quad \quad \quad \; Output\\
    \midrule
    $a$ \quad \quad \quad \ \textcolor{cyan}{$Q_1A_1\cdots Q_{n-1}A_{n-1}$}$\ <sep>\ answer\ this:\ $\textcolor{blue}{$Q_n$}\ $ <sep>\ $ \textcolor{cyan}{$S$} \quad \quad \quad \quad \quad \ \ \textcolor{teal}{$A_n$}  \\
    $q$ \quad \quad \quad \ \textcolor{cyan}{$Q_1A_1\cdots Q_{n-1}A_{n-1}$}$\ <sep>\ question\ it:\ $\textcolor{blue}{$A_n$} $<sep>\ $ \textcolor{cyan}{$S$} \quad \quad \quad \quad \quad \quad \ \textcolor{teal}{$Q_n$}  \\
    $main$ \quad \ \ \textcolor{cyan}{$Q_1A_1\cdots Q_{n-1}A_{n-1}$}$\ <sep>\ pose\ pair:\ $\textcolor{blue}{$R_n$} $ \ <sep>\ $ \textcolor{cyan}{$S$} \quad \quad \quad \quad \quad \quad \; \textcolor{teal}{$Q_n?A_n$}  \\
    $r$ \quad \quad \quad \ \textcolor{cyan}{$Q_1A_1\cdots Q_{n-1}A_{n-1}$}$\ <sep>\ find\ rationale:\ $ \textcolor{blue}{$Q_{n}A_{n}$}$ <sep>\ $ \textcolor{cyan}{$S$} \quad \quad \quad \ \textcolor{teal}{$R_n$}  \\
    $h$ \quad \quad \quad \ \textcolor{cyan}{$Q_1A_1\cdots$} \textcolor{blue}{$Q_{n}A_{n}$}$\ <sep>\ generate\ history<sep>$ \quad \quad \quad \quad \quad \quad \quad \quad \; \textcolor{teal}{$\bigcup_{i=1}^{n}R_i$}  \\
    \bottomrule
    \end{tabular}
    \caption{An ideal CQG example, where the questions are mutually connected and can cover diverse information to help understand the whole story. Also an example of data composition of our multitask generation framework, as well as the input and output in the $n^{th}$ generation step. In this example, the output of \textit{Task} $h$ is $stc_1$ when $n=1$, and is $stc_1stc_2$ when $n\ge 2$. ``$\bigcup$'' means coverage, or union set, with no overlap or replication.}
    \label{tb2}
\vspace{-0.3cm}
\end{table*}

Finally, we conduct abundant experiments to augment various QA datasets, only using the model trained on CoQA. We also make a manual evaluation and propose a novel zero-shot method for document-level NLI task \citep{yin-etal-2021-docnli} using question generation. Successfully, we promote the performance on multiple QA scenes and prove the expansibility of our model on different NLP tasks.


\section{Related Work}
%
%
Question generation is a promising task which has been well studied in many researches. Initially, rule-based or traditional machine learning methods are widely used in producing questions. \citet{heilman-smith-2010-good} adopt verb transformations and
\citet{berant-etal-2013-semantic} use semantic parsing to synthesize questions. Recently, deep learning techniques have given a further development of question generation. \citet{du-etal-2017-learning} use an LSTM \citep{hochreiter-etal-1997-lstm} model, and \citet{sultan-etal-2020-importance} adopt RoBERTa \citep{liu-etal-2019-roberta} model to generate questions.

At the same time, the strategies like multitask learning and self-training have been applied to improve the quality of generated questions. \citet{zhou-etal-2019-multi} and \citet{ma-etal-2020-improving} employ a multitask structure to generate coherent and fluent questions. \citet{sachan-xing-2018-self} and \citet{rennie-etal-2020-unsupervised} adopt a self-training strategy to jointly learn to ask and answer questions. \citet{alberti-etal-2019-synthetic} use roundtrip consistency to filter out inconsistent results. \citet{shinoda-etal-2021-improving} generate noisy data and \citet{sultan-etal-2020-importance} employ nucleus sampling \citep{ari-etal-2020-curious} to improve the diversity of questions. However, they mainly focus on only one quality aspect and most of them are based on pre-defined answers or original data.

As QG can produce meaningful questions, it has been widely used to promote other NLP tasks. \citet{liu-etal-2020-tell} use a constrained question rewriting way to generate new data for QA tasks. \citet{wang-etal-2020-asking} and \citet{nan-etal-2021-improving} check the faithfulness of summaries through answering generated questions. \citet{pan-etal-2021-qacg} generate question-answer pairs and convert them for fact verification. Nevertheless, the researches above mainly produce each question independently and ignore the connections between questions.

As for generating a set of questions over a specific passage, \citet{krishna-iyyer-2019-generating} propose a pipelined system to ask different levels of questions from general to specific. \citet{lee-etal-2020-generating} use conditional variational autoencoder to generate multiple robust questions for a given paragraph. Similar to us, \citet{chai-2020-learning} generate sequential and related questions under dual-graph interaction, but use ground-truth answers.
To the best of our knowledge, we are the first to consecutively synthesize a series of connected question-answer pairs to understand an entire passage, with the comprehensive consideration of accuracy, coverage, and informativeness.

\section{Multitask Framework}
In our CQG strategy, the foundation is five various but unified tasks. The effects of these tasks are dynamically spread throughout our whole strategy. In section \ref{sec4} we use them to compose four related methods to enhance different stages.

We first symbolically define the four key elements used in our work. $S$ denotes the story from which questions are produced; $Q_n$ means the $n^{th}$ question and $A_n$ is the answer; $R_n$ is the corresponding rationale (always one sentence) based on which $Q_n$ is generated. Since the Q-A pairs are generated dependently on previous questions, $C_n$ denotes the context which composes of previous $n-1$ Q-A pairs and the story.\footnote{Please be aware that story is the text content, and context is story plus previous $n-1$ Q-A pairs.}
Table \ref{tb2} is an example. Then we define the main task and the four auxiliary tasks using the $n^{th}$ turn as follows:\\
\textit{Task} $main$: $C_n+R_n\rightarrow Q_n+A_n$\\
\textit{Task} $a$: $C_n+Q_n\rightarrow A_n$ \\
\textit{Task} $q$: $C_n+A_n\rightarrow Q_n$ \\
\textit{Task} $r$: $C_n+Q_n+A_n\rightarrow R_n$ \\
\textit{Task} $h$: $\sum_{i=1}^{n}(Q_i+A_i)\rightarrow \bigcup_{i=1}^{n}R_i$


In \textit{Task} $main$, because we think the extractive answer is usually simple and it is inconsistent to get a Q-A in two steps, different from traditional methods, we input the context and rationale and output the question and answer simultaneously. 

The design of \textit{Task} $a$ and \textit{Task} $q$ aims to guarantee that the generated question and answer are accurate: given the question we can get the answer and given the answer we can get the question. Here \textit{Task} $a$ follows traditional QA form. We do not input the rationale in \textit{Task} $q$ because previous Q-A pairs are included in the context, so if $A_n$ is an accurate answer, the model should recognize the connection between the answer and the previous Q-A pairs, and restore the question easily.



Moreover, although we input the rationale in \textit{Task} $main$, it does not necessarily imply that the question-answer pair is derived from it. So we design \textit{Task} $r$ ($ C_n+Q_n+A_n\rightarrow R_n$) to verify that the model indeed uses the information in input rationale to get the question and answer. \textit{Task} $r$ helps the model to recognize the corresponding rationale, and then increase the coverage of a Q-A series, which means more events or more segments are precisely referred to.

Finally, to generate an informative and useful question, which means the knowledge it asks for does not overlap with previous ones, we consider that the more unseen information included in the Q-A pair, the better. We introduce the history of the context as the coverage of all previous rationales, which represents the total background information till the current Q-A turn. Therewith, we present \textit{Task} $h$: $\sum_{i=1}^{n}(Q_i+A_i)\rightarrow \bigcup_{i=1}^{n}R_i$, which uses Q-A pairs to restore the history. ``$\bigcup$'' means cover, with no overlap or replication, and ``+'' means append or plus.



Both \textit{Task} $r$ and \textit{Task} $h$ use Q-A pairs to restore the context, but focus on coverage and informativeness differently. Specifically, a part of a story is covered means a question is asked based on it, but a informative question means it is non-trivial and important and contains no repetitive information. Also, in \textit{Task} $r$ we input the context, so the model only needs to locate the correct rationale, but in \textit{Task} $h$, it has to generate the history completely based on Q-A pairs. Therefore in \textit{Task} $h$, if the $n^{th}$ Q-A pair carries more unseen information, it will be easier to restore the history compared with a Q-A pair with repetitive or trivial information.

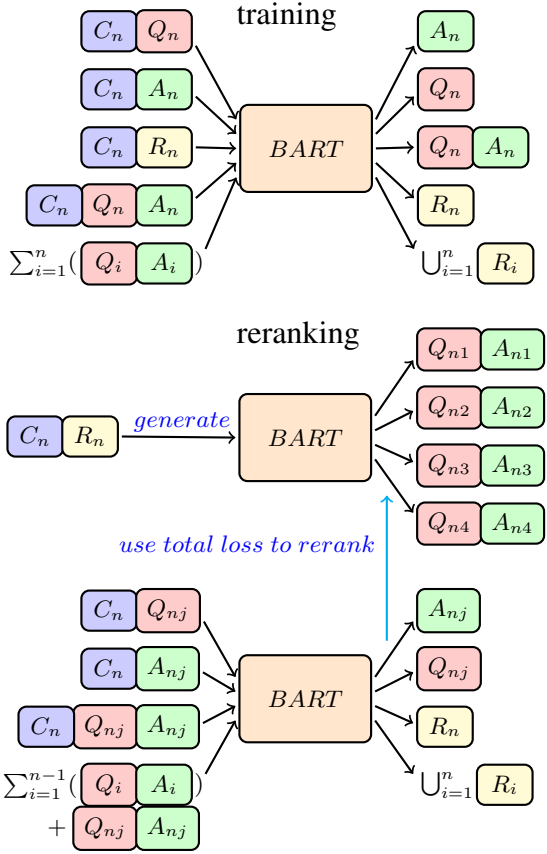
\begin{figure}[!t]
    \tikzstyle{BGnode} = [rounded corners=0pt,inner sep=4pt,minimum height=20em,minimum width=20em,fill=cyan!0]
    \tikzstyle{BG2node} = [rounded corners=0pt,inner sep=4pt,minimum height=19.5em,minimum width=20em,fill=green!0]
    \tikzstyle{Cnode} = [rounded corners=3pt,inner sep=4pt,minimum height=1.3em,minimum width=1.9em,draw,thick,fill=blue!20]
    \tikzstyle{Qnode} = [rounded corners=3pt,inner sep=4pt,minimum height=1.3em,minimum width=1.9em,draw,thick,fill=red!20]
    \tikzstyle{Anode} = [rounded corners=3pt,inner sep=4pt,minimum height=1.3em,minimum width=1.9em,draw,thick,fill=green!20]
    \tikzstyle{Snode} = [rounded corners=3pt,inner sep=4pt,minimum height=1.3em,minimum width=1.9em,draw,thick,fill=yellow!20]
    \tikzstyle{Hnode} = [rounded corners=3pt,inner sep=4pt,minimum height=1.3em,minimum width=1.9em,draw,thick,fill=black!20]
    \tikzstyle{Bart} = [rounded corners=3pt,inner sep=4pt,minimum height=3em,minimum width=4.5em,draw,thick,fill=orange!20]
    \begin{tikzpicture}
        \node [Bart,anchor=west,draw=black!100] (bart0) at (0,0) {\small{$BART$}};
        \node [anchor=west,black] (training) at ([shift={(-5em,3em)}]bart0.north east) {\large{training}};
        \node [Cnode,anchor=west] (c1l1) at ([shift={(-10em,2.5em)}]bart0.north east) {\small{$C_n$}};
        \node [Qnode,anchor=west] (q1l1) at ([shift={(-0.1em,0em)}]c1l1.east) {\small{$Q_n$}};
        \node [Cnode,anchor=west] (c1l2) at ([shift={(-10em,0.5em)}]bart0.north east) {\small{$C_n$}};
        \node [Anode,anchor=west] (a1l2) at ([shift={(-0.1em,0em)}]c1l2.east) {\small{$A_n$}};
        \node [Cnode,anchor=west] (c1l3) at ([shift={(-10em,-1.5em)}]bart0.north east) {\small{$C_n$}};
        \node [Snode,anchor=west] (s1l3) at ([shift={(-0.1em,0em)}]c1l3.east) {\small{$R_n$}};
        \node [Qnode,anchor=west] (q1l4) at ([shift={(-10em,-3.5em)}]bart0.north east) {\small{$Q_n$}};
        \node [Cnode,anchor=east] (c1l4) at ([shift={(0.1em,0em)}]q1l4.west) {\small{$C_n$}};
        \node [Anode,anchor=west] (a1l4) at ([shift={(-0.1em,0em)}]q1l4.east) {\small{$A_n$}};
        \node [Qnode,anchor=west] (q1l5) at ([shift={(-10em,-5.5em)}]bart0.north east) {\small{$Q_i$}};
        \node [Anode,anchor=west] (a1l5) at ([shift={(-0.1em,0em)}]q1l5.east) {\small{$A_i$}};
        \node [anchor=west] (l1l5) at ([xshift=-2.8em]q1l5.west) {\small{$\sum_{i=1}^{n}($}};
        \node [anchor=west] (r1l5) at ([xshift=-0.3em]a1l5.east) {\small{$)$}};
        
        \draw [->,thick] ([xshift=0.1em,yshift=-0.5em]q1l1.east) -> ([xshift=-0.1em,yshift=1.0em]bart0.west);
        \draw [->,thick] ([xshift=0.1em,yshift=-0.25em]a1l2.east) -> ([xshift=-0.1em,yshift=0.5em]bart0.west);
        \draw [->,thick] ([xshift=0.1em,yshift=0em]s1l3.east) -> ([xshift=-0.1em,yshift=0em]bart0.west);
        \draw [->,thick] ([xshift=0.1em,yshift=0.25em]a1l4.east) -> ([xshift=-0.1em,yshift=-0.5em]bart0.west);
        \draw [->,thick] ([xshift=-0.3em,yshift=0.5em]r1l5.east) -> ([xshift=-0.1em,yshift=-1.0em]bart0.west);
        
        \node [Anode,anchor=west] (a1r1) at ([shift={(1.5em,2.5em)}]bart0.north east) {\small{$A_n$}};
        \node [Qnode,anchor=west] (q1r2) at ([shift={(1.5em,0.5em)}]bart0.north east){\small{$Q_n$}};
        \node [Qnode,anchor=west] (q1r3) at ([shift={(1.5em,-1.5em)}]bart0.north east) {\small{$Q_n$}};
        \node [Anode,anchor=west] (a1r3) at ([xshift=-0.1em]q1r3.east){\small{$A_n$}};
        \node [Snode,anchor=west] (s1r4) at ([shift={(1.5em,-3.5em)}]bart0.north east) {\small{$R_n$}};
        \node [anchor=west] (l1r5) at ([shift={(1.3em,-5.5em)}]bart0.north east) {\small{$\bigcup_{i=1}^{n}$}};
        \node [Snode,anchor=west] (s1r5) at ([xshift=-0.3em]l1r5.east) {\small{$R_i$}};
        
        \draw [->,thick] ([xshift=0.1em,yshift=1.0em]bart0.east) -> ([shift={(-0.1em,-0.5em)}]a1r1.west);
        \draw [->,thick] ([xshift=0.1em,yshift=0.5em]bart0.east) -> ([shift={(-0.1em,-0.25em)}]q1r2.west);
        \draw [->,thick] ([xshift=0.1em,yshift=0em]bart0.east) -> ([shift={(-0.1em,0em)}]q1r3.west);
        \draw [->,thick] ([xshift=0.1em,yshift=-0.5em]bart0.east) -> ([shift={(-0.1em,0.25em)}]s1r4.west);
        \draw [->,thick] ([xshift=0.1em,yshift=-1.0em]bart0.east) -> ([shift={(0.1em,0.5em)}]l1r5.west);

        \node [anchor=west,black] (infering) at ([shift={(-5em,-8em)}]bart0.north east) {\large{reranking}};
        \node [Bart,anchor=west,draw=black!100] (bart1) at ([shift={(-4.6em,-11.5em)}]bart0.north east) {\small{$BART$}};
        \node [Cnode,anchor=west] (c2l1) at ([shift={(-12.5em,-1.5em)}]bart1.north east) {\small{$C_n$}};
        \node [Snode,anchor=west] (s2l1) at ([shift={(-0.1em,0em)}]c2l1.east) {\small{$R_n$}};
        \draw [->,thick] ([xshift=0.1em]s2l1.east) -> ([xshift=-0.1em,yshift=0em]bart1.west);
        \node [anchor=west,blue] (generate) at ([shift={(-8.5em,-1em)}]bart1.north east) {\small{$generate$}};
        
        \node [Qnode,anchor=west] (q2r1) at ([shift={(1.5em,1.5em)}]bart1.north east) {\small{$Q_{n1}$}};
        \node [Anode,anchor=west] (a2r1) at ([shift={(-0.1em,0em)}]q2r1.east) {\small{$A_{n1}$}};
        \node [Qnode,anchor=west] (q2r2) at ([shift={(1.5em,-0.5em)}]bart1.north east) {\small{$Q_{n2}$}};
        \node [Anode,anchor=west] (a2r2) at ([shift={(-0.1em,0em)}]q2r2.east) {\small{$A_{n2}$}};
        \node [Qnode,anchor=west] (q2r3) at ([shift={(1.5em,-2.5em)}]bart1.north east) {\small{$Q_{n3}$}};
        \node [Anode,anchor=west] (a2r3) at ([shift={(-0.1em,0em)}]q2r3.east) {\small{$A_{n3}$}};
        \node [Qnode,anchor=west] (q2r4) at ([shift={(1.5em,-4.5em)}]bart1.north east) {\small{$Q_{n4}$}};
        \node [Anode,anchor=west] (a2r4) at ([shift={(-0.1em,0em)}]q2r4.east) {\small{$A_{n4}$}};
        
        \draw [->,thick] ([xshift=0.1em,yshift=0.75em]bart1.east) -> ([shift={(-0.1em,-0.375em)}]q2r1.west);
        \draw [->,thick] ([xshift=0.1em,yshift=0.25em]bart1.east) -> ([shift={(-0.1em,-0.125em)}]q2r2.west);
        \draw [->,thick] ([xshift=0.1em,yshift=-0.25em]bart1.east) -> ([shift={(-0.1em,0.125em)}]q2r3.west);
        \draw [->,thick] ([xshift=0.1em,yshift=-0.75em]bart1.east) -> ([shift={(-0.1em,0.375em)}]q2r4.west);

        \node [Bart,anchor=west,draw=black!100] (bart2) at ([shift={(-4.6em,-20.5em)}]bart0.north east) {\small{$BART$}};
        \node [Cnode,anchor=east] (c3l1) at ([shift={(-8em,1.5em)}]bart2.north east) {\small{$C_n$}};
        \node [Qnode,anchor=west] (q3l1) at ([shift={(-0.1em,0em)}]c3l1.east) {\small{$Q_{nj}$}};
        \node [Cnode,anchor=east] (c3l2) at ([shift={(-8em,-0.5em)}]bart2.north east) {\small{$C_n$}};
        \node [Anode,anchor=west] (a3l2) at ([shift={(-0.1em,0em)}]c3l2.east) {\small{$A_{nj}$}};
        \node [Qnode,anchor=east] (q3l3) at ([shift={(-8em,-2.5em)}]bart2.north east) {\small{$Q_{nj}$}};
        \node [Cnode,anchor=east] (c3l3) at ([shift={(0.1em,0em)}]q3l3.west) {\small{$C_n$}};
        \node [Anode,anchor=west] (a3l3) at ([shift={(-0.1em,0em)}]q3l3.east) {\small{$A_{nj}$}};
        \node [Qnode,anchor=east] (q3l4) at ([shift={(-8em,-6.0em)}]bart2.north east) {\small{$Q_{nj}$}};
        \node [Anode,anchor=west] (a3l4) at ([shift={(-0.1em,0em)}]q3l4.east) {\small{$A_{nj}$}};
        \node [Qnode,anchor=east] (q3l5) at ([shift={(-8em,-4.5em)}]bart2.north east) {\small{$Q_i$}};
        \node [Anode,anchor=west] (a3l5) at ([shift={(-0.1em,0em)}]q3l5.east) {\small{$A_i$}};
        \node [anchor=west] (l3l5) at ([xshift=-3.0em]q3l5.west) {\small{$\sum_{i=1}^{n-1}($}};
        \node [anchor=west] (r3l5) at ([xshift=-0.3em]a3l5.east) {\small{$)$}};
        \node [anchor=west] (jia) at ([xshift=-1.2em]q3l4.west) {\small{$+$}};
        
        \draw [->,thick] ([xshift=0.1em,yshift=-0.375em]q3l1.east) -> ([xshift=-0.1em,yshift=0.75em]bart2.west);
        \draw [->,thick] ([xshift=0.1em,yshift=-0.125em]a3l2.east) -> ([xshift=-0.1em,yshift=0.25em]bart2.west);
        \draw [->,thick] ([xshift=0.1em,yshift=0.125em]a3l3.east) -> ([xshift=-0.1em,yshift=-0.25em]bart2.west);
        \draw [->,thick] ([xshift=-0.3em,yshift=0.375em]r3l5.east) -> ([xshift=-0.1em,yshift=-0.75em]bart2.west);
        
        \node [Anode,anchor=west] (a3r1) at ([shift={(1.5em,1.5em)}]bart2.north east) {\small{$A_{nj}$}};
        \node [Qnode,anchor=west] (q3r2) at ([shift={(1.5em,-0.5em)}]bart2.north east){\small{$Q_{nj}$}};
        \node [Snode,anchor=west] (s3r3) at ([shift={(1.5em,-2.5em)}]bart2.north east) {\small{$R_{n}$}};
        \node [anchor=west] (l3r4) at ([shift={(1.3em,-4.5em)}]bart2.north east) {\small{$\bigcup_{i=1}^{n}$}};
        \node [Snode,anchor=west] (s3r4) at ([xshift=-0.3em]l3r4.east) {\small{$R_i$}};
        
        \draw [->,thick] ([xshift=0.1em,yshift=0.75em]bart2.east) -> ([shift={(-0.1em,-0.375em)}]a3r1.west);
        \draw [->,thick] ([xshift=0.1em,yshift=0.25em]bart2.east) -> ([shift={(-0.1em,-0.125em)}]q3r2.west);
        \draw [->,thick] ([xshift=0.1em,yshift=-0.25em]bart2.east) -> ([shift={(-0.1em,0.125em)}]s3r3.west);
        \draw [->,thick] ([xshift=0.1em,yshift=-0.75em]bart2.east) -> ([shift={(0.1em,0.375em)}]l3r4.west);
        
        \draw [->,thick,cyan] ([xshift=0.5em,yshift=2em]bart2.east) -> ([xshift=0.5em,yshift=-2em]bart1.east);
        \node [anchor=west,blue] (rerank) at ([shift={(-9.0em,3.8em)}]bart2.north east) {\small{$use \; total \; loss \; to \; rerank$}};
        
    \end{tikzpicture}
    \caption{An overview of our dynamic multitask framework during joint training and self-reranking. One main task generates Q-A pairs and four auxiliary tasks generate other four CQG elements. In training, the five tasks are jointly trained in one model. In inference, the model uses the main task to generate candidates and then uses the auxiliary tasks to self-rerank them. We use the $n^{th}$ turn of a series of questions as an example and generate 4 candidates in inference. $j \in \{1,2,3,4\}$.}
    \label{fg1}
\vspace{-0.5cm}
\end{figure}

\section{Training and Inference}
\label{sec4}
Based on the dynamic multitask framework, we jointly train a BART \citep{lewis-etal-2020-bart} model.
In inference, we use the main task to generate several candidates and self-rerank them using the auxiliary tasks. With the reranking losses, we design a formula to assess the information and automatically sample the rationales. Globally, we beam-search for the best Q-A series on sentence level.

\subsection{Joint Training}
We randomly shuffle the five kinds of training instances and use a BART model to jointly train the five tasks together. We also train the model to generate a ``?'' between a Q-A to split it, and adopt five hand-made prompts \citep{liu-etal-2021-prompt}. Table \ref{tb2} shows an example of our data structure. Given the Seq2Seq model parameterized by $\theta$, the input sequence $\bm{x}$ with $n$ tokens = $\{x_1,\cdots,x_n\}$ and label $\bm{y}$ with $m$ tokens = $\{y_1,\cdots,y_m\}$, the generation probability and loss are as follows:
\begin{equation}
    p(\bm{y}|\bm{x},\theta)=\prod_{z=1}^{m}p(\bm{y}_z|\bm{y}_{<z},\bm{x},\theta)\label{eq1}
\end{equation}
\begin{equation}
    loss(\bm{y}|\bm{x},\theta)=-\frac{1}{m}\sum_{z=1}^{m}log\,p(\bm{y}_z|\bm{y}_{<z},\bm{x},\theta)\label{eq2}
\end{equation}

Through joint training we train a model to learn from different views and allow every task to benefit each other mutually. We also acquire the ability to do all five tasks in one model.

\subsection{Self-Reranking}
\label{m32}
During the inference stage, through the main task we can obtain many candidate question-answer pairs using a decoding strategy like nucleus sampling. To select the best result, inspired by \citet{shen-etal-2021-generate-rank}, we employ these candidates to the same model to do \textit{Task} $a$,$q$,$r$, and $h$, and then rank the candidates using the inference losses of the four auxiliary tasks. In another word, we use one model as both the generator and ranker. During reranking, the corresponding question and answer of the auxiliary tasks are those generated from \textit{Task} $main$. Specifically, we multiply the four losses together as the reranking loss, as Eq.\ref{eq3}, where the subscript $i$ refers to different tasks. We also design other loss aggregation methods to calculate the reranking losses, as in Appendix \ref{apbrl}, which shows that using $\prod$ or $\sum$ are the same in nature.
\begin{equation}
    loss_{rank}(\bm{y}|\bm{x},\theta)=\prod_{i\in \{a,q,r,h\}}loss(\bm{y}_i|\bm{x}_i,\theta)\label{eq3}
\end{equation}

We consider the candidate with the lowest reranking loss as the one who excels in accuracy, coverage, and informativeness generally. This is inspired by the idea of evaluating generated text as text generation \citep{yuan-etal-2021-bartscore}. Through this strategy we also unify the form of training and reranking process and manage to do them in the same model. Figure \ref{fg1} shows the structure of our multitask joint training and self-reranking.

\subsection{Rationale Sampling}
\label{m33}
The aforementioned methods are useful to generate one good Q-A pair. Still, how to effectively generate consecutive questions on a passage remains unsettled. By default, we select every rationale as the next sentence of previous one. However, one rationale does not necessarily correspond to only one question, because a long informative sentence may be suitable for several Q-A pairs.

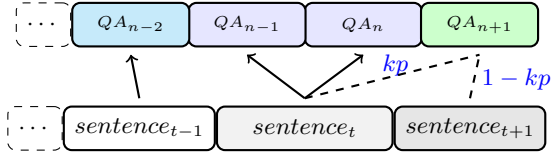
\begin{figure}[!t]
    \centering
    \begin{tikzpicture}
        
        \tikzstyle{BGnode} = [rounded corners=0pt,inner sep=4pt,minimum height=6em,minimum width=20em,fill=cyan!0]
        \tikzstyle{leftnode} = [rounded corners=3pt,inner sep=4pt,minimum height=1.5em,minimum width=5em,draw,thick,fill=black!0]
        \tikzstyle{midnode} = [rounded corners=3pt,inner sep=4pt,minimum height=1.5em,minimum width=6em,draw,thick,fill=black!5]
        \tikzstyle{rightnode} = [rounded corners=3pt,inner sep=4pt,minimum height=1.5em,minimum width=5em,draw,thick,fill=black!10]
        \tikzstyle{midqanode} = [rounded corners=3pt,inner sep=4pt,minimum height=1.5em,minimum width=4em,draw,thick,fill=blue!10]
        \tikzstyle{leftqanode} = [rounded corners=3pt,inner sep=4pt,minimum height=1.5em,minimum width=4em,draw,thick,fill=cyan!20]
        \tikzstyle{rightqanode} = [rounded corners=3pt,inner sep=4pt,minimum height=1.5em,minimum width=4em,draw,thick,fill=green!20]
        \tikzstyle{blackqanode} = [rounded corners=3pt,inner sep=4pt,minimum height=1.5em,minimum width=1em,draw,dashed,fill=black!0]
        
        \node [midnode,anchor=west,draw=black!100] (mids) at (0,0) {\small{$sentence_t$}};
        \node [rightnode,anchor=west,draw=black!100] (rights) at ([shift={(0em,0em)}]mids.east) {\small{$sentence_{t+1}$}};
        \node [leftnode,anchor=east,draw=black!100] (lefts) at ([shift={(0em,0em)}]mids.west) {\small{$sentence_{t-1}$}};
        \node [blackqanode,anchor=east,draw=black!100] (llefts) at ([shift={(0em,0em)}]lefts.west) {\small{$\cdots$}};;
        
        \node [midqanode,anchor=south,draw=black!100] (qam) at ([shift={(2em,2em)}]mids.north) {\tiny{$QA_n$}};
        \node [midqanode,anchor=east,draw=black!100] (qam-1) at ([shift={(0.1em,0em)}]qam.west) {\tiny{$QA_{n-1}$}};
        \node [leftqanode,anchor=east,draw=black!100] (qam-2) at ([shift={(0.1em,0em)}]qam-1.west) {\tiny{$QA_{n-2}$}};
        \node [rightqanode,anchor=west,draw=black!100] (qam+1) at ([shift={(-0.1em,0em)}]qam.east) {\tiny{$QA_{n+1}$}};
        \node [blackqanode,anchor=east,draw=black!100] (qam-3) at ([shift={(0em,0em)}]qam-2.west) {\small{$\cdots$}};
        
        \node [anchor=west,blue] (rp) at ([shift={(2.3em,1.4em)}]mids.north) {\small{$kp$}};
        \node [anchor=west,blue] (1-rp) at ([shift={(5.7em,0.9em)}]mids.north) {\small{$1-kp$}};
        
        \draw [->,thick] ([xshift=0em,yshift=0.25em]lefts.north) -> ([shift={(0em,-0.25em)}]qam-2.south);
        \draw [->,thick] ([xshift=0em,yshift=0.25em]mids.north) -> ([shift={(0em,-0.25em)}]qam-1.south);
        \draw [->,thick] ([xshift=0em,yshift=0.25em]mids.north) -> ([shift={(0em,-0.25em)}]qam.south);
        \draw [dashed, thick] ([xshift=0em,yshift=0.25em]mids.north) -> ([shift={(0em,-0.25em)}]qam+1.south);
        \draw [dashed, thick] ([xshift=0em,yshift=0.25em]rights.north) -> ([shift={(0em,-0.25em)}]qam+1.south);
    \end{tikzpicture}
    \caption{An example of rationale sampling, in which there is a probability of $kp$ that $R_{n+1}$ is $sentence_t$, and $1-kp$ it is $sentence_{t+1}$. Specifically in this example, $n'$ is $n-2$, $R_{n'}$ is $sentence_{t-1}$, and $m_{n'}$ is the length of $\sum_{i=1}^{t-1} sentence_i.$}
    \label{fg2}
    \vspace{-0.3cm}
\end{figure}

Hence, we propose the rationale sampling strategy, which introduces a probability that the next rationale keeps the same sentence as the current one, as Figure \ref{fg2} shows. We use $kp$ as the keeping probability. Then intuitively, we let $kp$ be linearly related to the amount of information left in the current rationale. Traditionally, the information is hard to be calculated quantitatively. However, recall that we use the loss of \textit{Task} $h$ to measure the information of a Q-A series, so similarly, we design a inference loss to represent the rest information in current rationale. We want a higher loss to mean that less information of $R_n$ is included in the Q-A series, and more information is still left in $R_n$.

Naturally, we first separate out the Q-A pairs on $R_n$. Given current step $n$, we find $n'$, which is the most recent step where $R_{n'} \neq R_n$. Then, we use 
{\setlength\abovedisplayskip{0cm}
\setlength\belowdisplayskip{0.25cm}
\begin{equation*}
\begin{split}
    &loss (R_n | \sum_{i=n'+1}^{n} (Q_i+A_i)+\bigcup_{i=1}^{n'}R_i, \theta)
    \\ &\approx \frac{m_nloss_{h_n}-m_{n'}loss_{h_{n'}}}{m_n-m_{n'}} \triangleq a
\end{split}
\end{equation*}}
to represent the rest information in $R_n$\footnote{$m_n$ = len($\bigcup_{i=1}^{n}R_i$). The details are in Appendix \ref{apbmrs}}, which is the loss of using previous sentences and the Q-A pairs on $R_n$ to restore $R_n$. 
Given our multitask framework, we use the ready-calculated losses of \textit{Task} $h$ to approximate this loss, without introducing more computation and complexity.

The approximation is $a$. Particularly if $n$ is 1, $a$ is $loss_{h_1}$. Empirically, we set the slope to be 0.2 and set a bound of 0-0.75. Finally, we get Eq.(\ref{eq5}), and the average $kp$ is 0.32 in the experiments, resulting in about 1.3 questions from one sentence.

\begin{equation}
    kp=\left\{
    \begin{aligned}
    0, \quad & \ a\le0 \\
    0.2a, \quad & \ 0<a<3.75 \\ 
    0.75, \quad & \ a\ge3.75 \\
    \end{aligned}
    \right.\label{eq5}
\end{equation}

Besides, we also design other two rationale sampling strategies as in Appendix \ref{apbrs}, which shows that our strategy which bases on \textit{Task} $h$ to calculate information performs better than other hand-made probability formulas.

\subsection{Sentence-Level Beam-Search}
Although rationale sampling helps catch more information and improves flexibility, it brings about more uncertainty. The mutually dependent generation may also lead to deviation \citep{li-etal-2021-ditch}. Thus, it is crucial to guide the flow direction in every step and ensure the quality of the whole series.

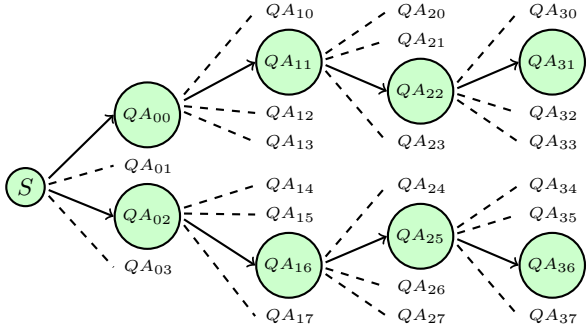
\begin{figure}[t]
    \tikzstyle{R} = [rounded corners=1pt,inner sep=2pt,circle,draw,thick,fill=green!20]
    \tikzstyle{BGnode} = [rounded corners=0pt,inner sep=4pt,minimum height=13em,minimum width=20em,fill=cyan!0]
    \begin{tikzpicture}
        \node [R,anchor=west,draw=black!100] (s) at (0,0) {\small{$S$}};
        
        \node [R,anchor=west,draw=black!100] (qa00) at ([shift={(3.5em,2.0em)}]s.north west) {\tiny{$QA_{00}$}};
        \node [anchor=west] (qa01) at ([shift={(3.5em,0.25em)}]s.north west) {\tiny{$QA_{01}$}};
        \node [R,anchor=west,draw=black!100] (qa02) at ([shift={(3.5em,-1.5em)}]s.north west) {\tiny{$QA_{02}$}};
        \node [anchor=west] (qa03) at ([shift={(3.5em,-3.25em)}]s.north west) {\tiny{$QA_{03}$}};
        \draw [->,thick] ([xshift=0.1em,yshift=0.3em]s.east) -> (qa00.west);
        \draw [dashed,thick] ([xshift=0.1em,yshift=0.1em]s.east) -> (qa01.west);
        \draw [->,thick] ([xshift=0.1em,yshift=-0.1em]s.east) -> (qa02.west);
        \draw [dashed,thick] ([xshift=0.1em,yshift=-0.3em]s.east) -> (qa03.west);
        
        \node [anchor=west] (qa10) at ([shift={(4.5em,2.75em)}]qa00.north west) {\tiny{$QA_{10}$}};
        \node [R,anchor=west,draw=black!100] (qa11) at ([shift={(4.5em,1.0em)}]qa00.north west) {\tiny{$QA_{11}$}};
        \node [anchor=west] (qa12) at ([shift={(4.5em,-0.75em)}]qa00.north west) {\tiny{$QA_{12}$}};
        \node [anchor=west] (qa13) at ([shift={(4.5em,-1.75em)}]qa00.north west) {\tiny{$QA_{13}$}};
        \draw [dashed,thick] ([xshift=0.1em,yshift=0.7em]qa00.east) -> (qa10.west);
        \draw [->,thick] ([xshift=0.1em,yshift=0.5em]qa00.east) -> (qa11.west);
        \draw [dashed,thick] ([xshift=0.1em,yshift=0.3em]qa00.east) -> (qa12.west);
        \draw [dashed,thick] ([xshift=0.1em,yshift=0.1em]qa00.east) -> (qa13.west);
        
        \node [anchor=west] (qa14) at ([shift={(4.5em,0.25em)}]qa02.north west) {\tiny{$QA_{14}$}};
        \node [anchor=west] (qa15) at ([shift={(4.5em,-0.75em)}]qa02.north west) {\tiny{$QA_{15}$}};
        \node [R,anchor=west,draw=black!100] (qa16) at ([shift={(4.5em,-2.5em)}]qa02.north west) {\tiny{$QA_{16}$}};
        \node [anchor=west] (qa17) at ([shift={(4.5em,-4.25em)}]qa02.north west) {\tiny{$QA_{17}$}};
        \draw [dashed,thick] ([xshift=0.1em,yshift=0.3em]qa02.east) -> (qa14.west);
        \draw [dashed,thick] ([xshift=0.1em,yshift=0.1em]qa02.east) -> (qa15.west);
        \draw [->,thick] ([xshift=0.1em,yshift=-0.1em]qa02.east) -> (qa16.west);
        \draw [dashed,thick] ([xshift=0.1em,yshift=-0.3em]qa02.east) -> (qa17.west);
        
        \node [anchor=west] (qa20) at ([shift={(4.5em,0em)}]qa10.west) {\tiny{$QA_{20}$}};
        \node [anchor=west] (qa21) at ([shift={(4.5em,0.75em)}]qa11.west) {\tiny{$QA_{21}$}};
        \node [R,anchor=west,draw=black!100] (qa22) at ([shift={(4.5em,0.75em)}]qa12.west) {\tiny{$QA_{22}$}};
        \node [anchor=west] (qa23) at ([shift={(4.5em,0em)}]qa13.west) {\tiny{$QA_{23}$}};
        \draw [dashed,thick] ([xshift=0.1em,yshift=0.3em]qa11.east) -> (qa20.west);
        \draw [dashed,thick] ([xshift=0.1em,yshift=0.1em]qa11.east) -> (qa21.west);
        \draw [->,thick] ([xshift=0.1em,yshift=-0.1em]qa11.east) -> (qa22.west);
        \draw [dashed,thick] ([xshift=0.1em,yshift=-0.3em]qa11.east) -> (qa23.west);
        
        \node [anchor=west] (qa24) at ([shift={(4.5em,0em)}]qa14.west) {\tiny{$QA_{24}$}};
        \node [R,anchor=west,draw=black!100] (qa25) at ([shift={(4.5em,-0.75em)}]qa15.west) {\tiny{$QA_{25}$}};
        \node [anchor=west] (qa26) at ([shift={(4.5em,-0.75em)}]qa16.west) {\tiny{$QA_{26}$}};
        \node [anchor=west] (qa27) at ([shift={(4.5em,0em)}]qa17.west) {\tiny{$QA_{27}$}};
        \draw [dashed,thick] ([xshift=0.1em,yshift=0.3em]qa16.east) -> (qa24.west);
        \draw [->,thick] ([xshift=0.1em,yshift=0.1em]qa16.east) -> (qa25.west);
        \draw [dashed,thick] ([xshift=0.1em,yshift=-0.1em]qa16.east) -> (qa26.west);
        \draw [dashed,thick] ([xshift=0.1em,yshift=-0.3em]qa16.east) -> (qa27.west);
        
        \node [anchor=west] (qa30) at ([shift={(9.0em,0em)}]qa10.west) {\tiny{$QA_{30}$}};
        \node [R,anchor=west,draw=black!100] (qa31) at ([shift={(9.0em,0em)}]qa11.west) {\tiny{$QA_{31}$}};
        \node [anchor=west] (qa32) at ([shift={(9.0em,0em)}]qa12.west) {\tiny{$QA_{32}$}};
        \node [anchor=west] (qa33) at ([shift={(9.0em,0em)}]qa13.west) {\tiny{$QA_{33}$}};
        \draw [dashed,thick] ([xshift=0.1em,yshift=0.3em]qa22.east) -> (qa30.west);
        \draw [->,thick] ([xshift=0.1em,yshift=0.1em]qa22.east) -> (qa31.west);
        \draw [dashed,thick] ([xshift=0.1em,yshift=-0.1em]qa22.east) -> (qa32.west);
        \draw [dashed,thick] ([xshift=0.1em,yshift=-0.3em]qa22.east) -> (qa33.west);
        
        \node [anchor=west] (qa34) at ([shift={(9.0em,0em)}]qa14.west) {\tiny{$QA_{34}$}};
        \node [anchor=west] (qa35) at ([shift={(9.0em,0em)}]qa15.west) {\tiny{$QA_{35}$}};
        \node [R,anchor=west,draw=black!100] (qa36) at ([shift={(9.0em,0em)}]qa16.west) {\tiny{$QA_{36}$}};
        \node [anchor=west] (qa37) at ([shift={(9.0em,0em)}]qa17.west) {\tiny{$QA_{37}$}};
        \draw [dashed,thick] ([xshift=0.1em,yshift=0.3em]qa25.east) -> (qa34.west);
        \draw [dashed,thick] ([xshift=0.1em,yshift=0.1em]qa25.east) -> (qa35.west);
        \draw [->,thick] ([xshift=0.1em,yshift=-0.1em]qa25.east) -> (qa36.west);
        \draw [dashed,thick] ([xshift=0.1em,yshift=-0.3em]qa25.east) -> (qa37.west);
        
        
    \end{tikzpicture}
    \caption{An overview of the sentence-level beam-search strategy. 
    In this example each step the model generates 4 question-answer candidates and the sentence-level beam size is 2.}
    \label{fg3}
\vspace{-0.4cm}
\end{figure}

Naturally, inspired by traditional beam-search (token-level), we propose the sentence-level beam-search, as Figure \ref{fg3} shows. Different from traditional beam-search, which generates a token in each search step, we generate a QA pair, and we adopt the reranking loss of each QA pair to take the place of the generation probability. Thus, in each step, we maintain several candidates with the lowest product of all previous reranking losses, which is calculated as Eq.\ref{eq6}, where $L$ is the final loss of our sentence-level beam-search method.
{\setlength\abovedisplayskip{0.25cm}
\setlength\belowdisplayskip{0.2cm}
\begin{equation}
\begin{aligned}
    L(Q_{1} A_{1}\cdots Q_{n} A_{n}|\bm{x},\theta)=\prod_{j=1}^{n}{loss_{rank}}_{_{j}}
\end{aligned}\label{eq6}
\end{equation}}

To summarize, 4.2 to 4.4 are for inference. Practically, in each generation step, we first use previous results to do rationale sampling to locate the rationale, then generate some candidates and calculate the current reranking losses, and finally we use the total losses to sentence-level beam-search and keep several Q-A flows for the next step.

\section{Experiments}
\subsection{Experimental Setup}
We employ CoQA \citep{reddy-etal-2019-coqa} training set as our training data. CoQA is a large-scale dataset for building Conversational Question Answering systems. The questions are conversational, and thus, every question after the first is dependent on the conversation history. The answers are free-form text with their corresponding rationales in the story. We expand the rationales to whole sentences and remove the questions with unknown answers. Finally, we get 7199 stories and each story has 15 turns of Q-A pairs on average. The training details and experiments are in Appendix \ref{apa}, where we also analyze the effect of joint training.

After training a model $\theta$ on CoQA, we evaluate our model by applying its question generation ability to two downstream tasks: data augmentation for QA and document-level NLI. Further, under the synthetic results on CoQA, we analyze their accuracy, coverage, and informativeness using human evaluations and a repeat-pose experiment.

\subsection{Experiments to Augment QA Data}
Data augmentation is one common way to employ generated questions and verify QG models. To augment QA dataset $D$, we (1) use $\theta$ to synthesize Q-A pairs $D'$ on the training set of $D$; (2) train another BART model $\theta'$ on $D'$ or $D+D'$ to answer questions\footnote{Since our synthetic Q-A pairs are free-form, we still use BART to generate the answers on both CoQA and SQuAD.}; (3) test $\theta'$ on the dev set of $D$. 

\subsubsection*{Results on CoQA}
First we test our strategy to augment CoQA dataset. The setting $Origin$ means the model $\theta'$ is trained on the original CoQA training set, and $Synth$ means it is trained with synthetic Q-A pairs. Inspired by \citet{yuan-etal-2021-bartscore}, we additionally use the inference losses to measure the performance

In $Synth$, we conduct single q, two step, and single m as three baseline models, where single q means we use a single \textit{Task} $q$ model to ask questions based on the origin answers, like the traditional QG methods. Two step means we first extract an answer\footnote{Use a BERT model to locate the start and end tokens.}, then generate a question on it using the single \textit{Task} $q$ model. Single m is a \textit{Task} $main$ model, which generates Q-A pairs.

Joint train is a multitask jointly trained model. Based on joint train model, we further add the self-reranking method, using all four auxiliary tasks. Then on this joint train + rerank model, we conduct four ablation studies of auxiliary tasks. 

Under joint train + rerank model, we also introduce other two conditions, independent and relay. By default, we generate the question series in an automatic way, which means every step the previous Q-A pairs are the Q-A pairs generated in previous steps. In independent condition, we let previous Q-A pairs be empty in all steps, which means the model generates every question like the first question, but when training QA model $\theta'$, we still input the previous QA pairs to align the data format with CoQA. In relay, the previous Q-A pairs of every synthetic instance are from CoQA training set, and the rationale is the ground-truth rationale sentence, which means the model inherits the Q-A flow from authentic CoQA's context.

Finally, still under joint train + rerank model, we add rationale sampling and sentence-level beam-search. Additionally, we merge the original training set with synthetic data to create the merging setting ($D+D'$). Note that RS and SBS are not suitable for independent or relay condition.

\begin{table}[!h]
    \renewcommand\tabcolsep{6pt}
    \renewcommand\arraystretch{1.4}
    \centering
    \small
    \begin{tabular}{lccc}
    \toprule
    \textbf{CoQA} & \textbf{Bleu} & \textbf{Infer Loss} & \textbf{F1$_{qa}$}\\
    \hline
    $Origin$ \\
    Bart & 38.52 & 0.777 & 78.54 \\
    $Synth$ \\
    Single q & 35.43/37.85  & 5.429/0.869 & 70.82/78.35\\
    Two step & 15.41/39.92  & 5.078/0.817 & 56.00/77.85 \\
    Single m & 27.04/41.42  & 5.538/0.776 & 65.66/79.20\\
    Joint train & 26.97/38.92 & 5.613/0.765 & 65.90/80.11 \\
    +rerank & 24.88/38.26 & 5.674/0.768 & 65.05/80.52 \\
    +RS & 31.73/46.24 & 5.323/\textbf{0.758}  & 72.33/81.83\\
    +SBS &32.01/\textbf{47.86} &5.431/0.766 &72.49/\textbf{81.98}\\
    \bottomrule
    \end{tabular}
    \caption{Results on CoQA dev set. In $Synth$, results without and with merging are separated by ``/''. In the middle are four ablation experiments of auxiliary tasks with Bart joint train+rerank. RS: rationale sampling. SBS: sentence-level beam-search.}
    \label{tb50}
\end{table}

\begin{table}[!h]
    \renewcommand\tabcolsep{6pt}
    \renewcommand\arraystretch{1.4}
    \centering
    \small
    \begin{tabular}{lccc}
    \toprule
    \textbf{CoQA} & \textbf{Bleu} & \textbf{Infer Loss} & \textbf{F1$_{qa}$}\\
    \hline
    Joint train & 26.97/38.92 & 5.613/0.765 & 65.90/80.11 \\
    +rerank a &25.31/38.03 &5.612/0.764 &63.71/80.23 \\
    +rerank q &24.66/37.83 &5.401/0.773 &64.44/80.29 \\
    +rerank r &24.03/38.05 &5.487/0.768 &63.73/80.18 \\
    +rerank h &23.10/37.32 &5.499/0.789 &63.01/80.27 \\
    +rerank all & 24.88/38.26 & 5.674/0.768 & 65.05/\textbf{80.52} \\
    \bottomrule
    \end{tabular}
    \caption{Results of ablation studies of four auxiliary tasks, on CoQA dev set.}
    \label{tb51}
\end{table}

\begin{table}[!h]
    \renewcommand\tabcolsep{2.5pt}
    \renewcommand\arraystretch{1.4}
    \centering
    \small
    \begin{tabular}{lccc}
    \toprule
    \textbf{CoQA} & \textbf{Bleu} & \textbf{Infer Loss} & \textbf{F1$_{qa}$}\\
    \hline
    Joint train + rerank & 24.88/38.26 & 5.674/0.768 & 65.05/80.52 \\
    indep & 20.38/39.03 & 5.490/0.783 & 56.54/78.29\\
    relay & 35.11/45.24  & 5.477/0.781 & 75.90/81.79\\
    +RS+SBS &32.01/47.86 &5.431/0.766 &72.49/\textbf{81.98}\\
    \bottomrule
    \end{tabular}
    \caption{Results of different conditions on CoQA dev set. ``+RS+SBS'' means Joint train+rerank+RS+SBS.}
    \label{tb52}
\end{table}

Table \ref{tb50} shows the main results. Table \ref{tb51} and Table \ref{tb52} are the results of ablation studies and different conditions. The single q and two step model make relatively low scores when merged with original data, which means they generate relatively simple and low-quality questions. Using our one step Q-A pairs generation, in merging setting the single m model leads to higher scores even than single q, which based on origin answers. Joint train and reranking further improve the F1$_{qa}$ scores by 1.32 points. From the four ablation studies in Table \ref{tb51}, it is not hard to see that every auxiliary task filters the results effectively, leading to 0.07 to 0.18 higher F1$_{qa}$ scores. 

As for our consecutive generation strategy, in Table \ref{tb52}, comparing the independent condition with our model, we can see that the consecutive generation largely improves the quality of questions by 2.23 F1$_{qa}$ scores. Moreover, although the relay model based on the original Q-A flow truly gets better performance, when we add RS and SBS strategy to get our best model, the F1$_{qa}$ score is further increased by 1.46 points, and finally it outperforms relay generation by 0.19 points. It shows that the Q-A series searched by RS and SBS are more proper even than the ground-truth flow.

\subsubsection*{Results on SQuAD and more data}
To check our QG ability on out-of-domain passages, we augment SQuAD \citep{rajpurkar-etal-2018-know} dataset using our best model trained on CoQA. We select the instances without unknown answers and with a story longer than 128 words. Since the questions in SQuAD are independent but also well-organized, we manually add previous Q-A pairs to align with CoQA.

To truly reveal the ability of our model, we employ it to synthesize more questions on a large number of unlabeled passages. We randomly collect 10000 Wikipedia passages whose lengths are from 100 to 500 words. Then we use our model trained on CoQA to generate questions on them, resulting in about 0.15 million Q-A pairs, which we use to augment both CoQA and SQuAD.
\begin{table}[!h]
    \renewcommand\tabcolsep{6pt}
    \renewcommand\arraystretch{1.4}
    \centering
    \small
    \begin{tabular}{lccc}
    \toprule
    \textbf{SQuAD} & \textbf{Bleu} & \textbf{Infer Loss} & \textbf{F1$_{qa}$}\\
    \hline
    $Origin$ \\
    Bart &65.52  &0.675  &84.26  \\
    +preQA &\textbf{68.67} &\textbf{0.625} &85.32 \\
    $Synth$ \\
    Ours & 41.91/67.43 & 4.639/0.691 & 67.57/85.59 \\
    +Wiki & 50.58/65.39  &4.010/0.630  &74.90/\textbf{85.88}  \\
    \bottomrule
    \textbf{CoQA} &  &  & \\
    \hline
    Ours &32.01/\textbf{47.86} &5.431/0.766 &72.49/81.98 \\
    +Wiki &33.01/47.43 &5.441/\textbf{0.758} &72.58/\textbf{82.21} \\
    Large &52.36 &0.521 &87.90 \\
    \bottomrule
    \end{tabular}
    \caption{Results of out-of-domain generation on SQuAD dev set, and on Wikipedia passages. ``Ours'' means Joint train+rerank+RS+SBS. ``Large'' means both the QG model and QA model are Bart Large, and the synthesized data for it is from CoQA and Wiki under the Joint train+rerank+RS+SBS setting. In $Synth$, results without and with merging are separated by ``/''.}
    \label{tb6}
     \vspace{-0.3cm}
\end{table}

Table \ref{tb6} shows the results. We can see that the Q-A series indeed enhances question answering. It also indicates that even if our model is trained on different dataset, its synthesized questions still help a QA model gain 0.27 more F1$_{qa}$ points on SQuAD. With more Wikipedia questions, in both CoQA and SQuAD, we manage to further improve F1$_{qa}$ by 0.29 and 0.23 scores. It shows that our model performs well when transferring to another dataset and can augment the QA training sets with large-scale unlabeled data. Finally we adopt $large$ model to get 87.90 F1$_{qa}$ points on CoQA.



\subsection{Understand a Whole Passage (DocNLI)}
To prove that our generated questions can really explore most information in an entire passage, we adopt our model for document-level NLI (DocNLI) task. Models are required to predict the relation (entailment or not) between a document-level premise and a hypothesis. 

Traditionally, a model predicts the relation in a sequence classification way. However, given our ability to synthesize consecutive questions to understand a passage, we propose a zero-shot method to predict the relation based on question generating and answering. Since entailment requires the hypothesis to be derived from the premise, we first generate Q-A pairs given the hypothesis, and then answer these questions based on the premise. If we can get the same answers, we predict entailment. In detail, we (1) use $\theta$ to synthesize a series of Q-A pairs on the hypothesis; (2) use $\theta$ to answer $Q$ on the premise, obtaining $A'$; (3) check the overlap (F1$_{qa}$) between $A$ and $A'$. If the F1$_{qa}$ exceeds a given threshold, it is entailment.

To make sure that the passages are long enough to generate a series of Q-A pairs, we select the instances whose premise and hypothesis are 200 to 1000 words from all train, dev, and test set of DocNLI, to be our evaluation set. It is 1677 instances in all, and we averagely generate 15 turns of Q-A each instance with rationale sampling. We use 60 points of F1$_{qa}$ as the threshold of entailment.

\begin{table}[!h]
    \renewcommand\tabcolsep{6pt}
    \renewcommand\arraystretch{1.4}
    \centering
    \small
    \begin{tabular}{lccc}
    \toprule
    \textbf{DocNLI} & \textbf{Infer Loss} & \textbf{F1$_{qa}$} & \textbf{F1$_{nli}$}\\
    \hline
    $Finetune$ \\
    Bert & - & - &48.56\\
    $QG$ \\
    Two step &1.142/2.020  &65.69/51.54  &47.67\\
    Single m &3.376/4.273  &61.00/47.73  &46.85\\
    Joint train &3.223/4.119  &63.32/49.56  &46.90\\
    \hdashline
    +rerank &3.217/4.149  &63.04/49.68  &47.91 \\
    indep &2.811/3.857  &63.90/49.18  &47.88\\
    \hdashline
    +RS &2.633/3.601 &65.98/50.99 &49.88 \\
    +SBS & 2.376/3.353 & 66.19/51.19 &\textbf{49.98}\\
    \bottomrule
    \end{tabular}
    \caption{Results of DocNLI task. Finetune is a BERT-base model fine-tuned on about 0.8 million other DocNLI instances. When using our zero-shot method, QA results of entailment and not entailment are separated by ``/''. We use different models for QG, and the QA model is the same as our best model $\theta$.}
    \label{tb8}
     \vspace{-0.3cm}
\end{table}

Tabel \ref{tb8} shows the results. F1$_{nli}$ is the harmonic mean of the precision and recall on the classification task. Impressively, using the zero-shot method, our best model surpasses the fine-tuned BERT model by 1.42 points of F1$_{nli}$ score. Among different QG settings, although two step model gets very low losses, its F1$_{nli}$ score is not very high, indicating that it generates relatively simple questions which cannot extract much information. Our one step model gets a lower F1$_{nli}$ score initially but with the joint training and reranking strategy, it improves the score by 0.98 points. Moreover, we can see clearly that the RS and SBS strategies improve the result significantly by 2.10 F1$_{nli}$ scores. They also manage to enlarge the discrimination between entailment and not entailment. It suggests that our consecutive generation strategy really produces question-answer pairs with most of the information in a passage, which can help understand the passage effectively.

\subsection{Analyses}
\subsubsection*{Accuracy and Coverage (\textit{Task} $a$, $q$ and $r$)}
Here we conduct two human evaluations, to prove that our strategy improves Q-A accuracy and story coverage, which are the effects of \textit{Task} $a$, $q$ and \textit{Task} $r$. Since the coverage requires the model to ask for more points of a passage, we use the question-rationale consistency (accuracy of rationale) to reflect it. This is because all sentences are asked at least once, and rationale sampling further guarantees the rationales to be well-distributed, so if the rationales are all precisely questioned, the coverage should be as well satisfactory.

We randomly collect 10\% stories from CoQA dev set and use different methods to generate Q-A pairs. We, the authors, then manually measure whether every question is correctly asked and answered and whether every question-answer pair is derived from its corresponding rationale.
\begin{table}[!h]
    \renewcommand\tabcolsep{6pt}
    \renewcommand\arraystretch{1.4}
    \centering
    \small
    \begin{tabular}{lcccc}
    \toprule
    \textbf{Acc of} & \textbf{Ours} & \textbf{-SBS} & \textbf{-Rerank} & \textbf{-Joint train} \\
    \hline
    Q-A pair &\textbf{94.85} &92.71 &90.32 &88.33 \\
    rationale &\textbf{95.65} &93.89 &90.97 &90.26 \\
    \bottomrule
    \end{tabular}
    \caption{Human evaluations of accuracy of Q-A and rationale. We do not ablate RS here because it is not relevant here and will make the data unaligned.}
    \label{tb9}
     \vspace{-0.3cm}
\end{table}

Table \ref{tb9} clearly shows that multitask joint training and reranking and sentence-level beam-search increase the accuracy of Q-A by 6.52 \% and rationale by 5.39 \%. Thus, we can say that our strategy, main due to \textit{Task} $a$, $q$ and \textit{Task} $r$, helps generate questions more correctly and locate the rationale more precisely, leading to higher Q-A accuracy and coverage in a series of questions.

\subsubsection*{Informativeness (\textit{Task} $h$)}
To evaluate the ability to utilize information in a rationale, we present the repeat-pose experiment on CoQA. It is adapted from relay condition, and requires the model to pose another question based on the same rationale and same context as the original question. In other words, the model has to ``squeeze'' more information from the same rationale, so the key is whether \textit{Task} $h$ can rank the informativeness of each candidate precisely.
\begin{table}[!h]
    \renewcommand\tabcolsep{4pt}
    \renewcommand\arraystretch{1.4}
    \centering
    \small
    \begin{tabular}{lccc}
    \toprule
    \textbf{CoQA} & \textbf{Bleu} & \textbf{Infer Loss} & \textbf{F1$_{qa}$}\\
    \hline
    Joint train relay w/o rerank & 41.01 & 0.737 & 81.21 \\
    Joint train repeat w/o rerank & 41.97 & 0.741 & 81.28 \\
    Joint train repeat w/ rerank & \textbf{43.40} & \textbf{0.708} & \textbf{81.57}\\
    \bottomrule
    \end{tabular}
    \caption{Results of the repeat-pose experiment. Synthetic data are merged with the original training set.}
    \label{tb10}
\end{table}

Table \ref{tb10} shows the results, which demonstrate that repeat-pose with self-reranking strategy further improves the F1$_{qa}$ scores by 0.36 points, indicating that \textit{Task} $h$ indeed helps select the more informative question-answer pairs.

\section{Conclusion}
In this paper, we propose the consecutive question generation task, which synthesizes mutually connected question-answer pairs to fully explore the information in a passage. By constructing a novel multitask framework with one main task and four unified auxiliary tasks, we generate optimum Q-A series using four sub-methods, which help ``generate good questions'' as well as ``find worth-asking information''. With extensive experiments, we prove that our model is able to generate high-quality Q-A pairs to understand a whole passage and has the power to benefit various NLP tasks.



\section*{Limitations}
In this paper, we propose a novel question generation strategy which can benefit multiple NLP scenes. For this work,
we summarize two limitations as follows.
First, CQG has high requirements for the training data. In this work, we adopt the CoQA corpus which is originally developed for the conversational QA task.
To the best of our knowledge, CoQA is the only existing dataset which is suitable for our task.
Without more datasets for evaluation,
we try to improve the performance on SQuAD and DocNLI to a certain degree by generating questions zero-shot or generating  questions on large-scale Wikipedia passages.
In future, we hope to build a CQG specific corpus and draw more attention to this novel task.

Second, the time cost of our strategy is higher than others', because we need to train five tasks jointly and rerank on four auxiliary tasks during inference. Specifically, it is about three times more in training and four times more in inference. Detailed analysis is in Appendix \ref{apb2}. In our future work, we will focus on the simplification of our strategy and the distillation of our model. Also, we will examine if a small model or a base model with fewer training data can get the same performance as other common models when using our strategy.

\section*{Acknowledgement}

We thank the anonymous reviewers for their helpful comments on this paper. This work was partially supported by National Key Research and Development Project
(2019YFB1704002) and National Natural Science
Foundation of China (61876009).
The corresponding author of this paper is Sujian Li.

\bibliography{anthology,custom}
\bibliographystyle{acl_natbib}

\appendix

\section{Implementation and Training Details}
\label{apa}
We use PyTorch to implement our models. We acquire the pre-trained BART model\footnote{\url{https://huggingface.co/facebook/bart-base}} from the Transformers library \citep{thomas-etal-2020-transformers}. 

During training, we set the batch size to 64 and learning rate to 1e-5. The maximum input length is 1024. In inference, we use beam-search with beam size 4 to generate answers for QA. Following \citet{sultan-etal-2020-importance}, we combine top-k sampling(k=50) with top-p sampling(p=0.95) to generate question-answer pairs. We averagely return 4 candidates each step and set sentence-level beam size to 4, which means in our best model, every step we select 4 out of 16 candidate Q-A flows. The models we use are $base$ size.

After training we evaluate the losses of five tasks on CoQA dev set, and the F1$_{qa}$ scores using \textit{Task} $a$. Table \ref{tb4} shows the results with different training settings. We can see that joint training improves the performance on four out of five tasks, suggesting that different tasks benefit each other effectively. Prompts also enhance the Q-A ability and decrease the losses on three out of five tasks. 

\begin{table}[!h]
    \renewcommand\tabcolsep{6pt}
    \renewcommand\arraystretch{1.4}
    \centering
    \small
    \begin{tabular}{lccc}
    \toprule
    \textbf{CoQA} & \textbf{Ours} & \textbf{w/o Prompts} & \textbf{w/o Joint}\\
    \hline
    $Loss$ a & \textbf{0.767} & 0.771 & 0.777\\
    $Loss$ q & \textbf{1.364} & 1.370 & 1.377\\
    $Loss$ m & \textbf{1.372} & 1.378 & 1.388\\
    $Loss$ r & 0.062 & \textbf{0.058} & 0.068\\
    $Loss$ h & 2.554 & 2.543 & \textbf{2.536}\\
    \midrule
    $F1_{qa}$ a & \textbf{80.60} & 80.07 & 78.54\\
    \bottomrule
    \end{tabular}
    \caption{Inference losses and F1$_{qa}$ scores on CoQA dev set using different training method.}
    \label{tb4}
\end{table}

During reranking, the scales of different losses are also not far from Table \ref{tb4}.

\section{Supplementary Analyses}
\subsection{Beam-Search or Nucleus Sampling}
\label{apb1}
As argued in \citep{sultan-etal-2020-importance}, nucleus sampling leads to higher diversity and is better than beam-search in QG. To verify that, we train two sets of models on different tasks with full strategies. We adopt beam-search with size 4 and nucleus sampling with top-k(k=50) and top-p(p=0.95). Table \ref{tb11} shows that nucleus sampling truly gains better results than beam-search.
\begin{table}[!h]
    \renewcommand\tabcolsep{6pt}
    \renewcommand\arraystretch{1.4}
    \centering
    \small
    \begin{tabular}{lcc}
    \toprule
    \textbf{Tasks} & \textbf{Beam-Search} & \textbf{Nucleus Sampling} \\
    \hline
    CoQA  &0.765/81.60 &0.766/\textbf{81.98} \\
    SQuAD  &0.679/85.51 &0.691/\textbf{85.59} \\ 
    DocNLI  & 2.380/49.33 & 2.376/\textbf{49.98} \\
    \bottomrule
    \end{tabular}
    \caption{Results using beam-search or nucleus sampling.}
    \label{tb11}
\end{table}

\subsection{Efficiency Analysis}
\label{apb2}
\begin{figure}[!h]
    \centering
    \begin{tikzpicture}[
        rednode/.style={shape=circle, draw=red, line width=3pt, inner sep=1pt,minimum size=3pt},
        bluenode/.style={shape=circle, draw=blue, line width=3pt, inner sep=1pt,minimum size=3pt},
        ]
        
        \matrix [draw,below left] at (7,7) {
          \node [rednode,label=right:\small{Single}] {}; \\
          \node [bluenode,label=right:\small{Multitask}] {}; \\
        };
        
        \matrix [draw,below left] at (7,11.6) {
          \node [rednode,label=right:\small{Single}] {}; \\
          \node [bluenode,label=right:\small{Multitask}] {}; \\
        };

        \draw[<->](7,3.9)--(0,3.9)--(0,7.5);
        \node[black] at (7.2,4.1) {\small{$step$}};
        \node[black] at (1,7.5) {\small{$loss \ of \ main$}};
        \node[red] at (3,4.7) {\small{$(13000,1.388)$}};
        \node[blue] at (6.4,4.5) {\small{$(55000,1.372)$}};
        \draw[red](0.0933,7.2471)--(0.1867,6.1866)--(0.2800,5.7690)--(0.3733,5.4696)--(0.4667,4.9569)--(0.5600,4.8039)--(0.6533,4.6977)--(0.7467,4.5909)--(0.8400,4.5309)--(0.9333,4.4556)--(1.0267,4.3953)--(1.1200,4.3611)--(1.2133,4.3239)--(1.3067,4.3122)--(1.4000,4.2651)--(1.4933,4.2429)--(1.5867,4.2567)--(1.6800,4.2300)--(1.7733,4.1949)--(1.8667,4.2069)--(1.9600,4.1937)--(2.0533,4.1781)--(2.1467,4.1511)--(2.2400,4.1838)--(2.3333,4.1718)--(2.4267,4.1508)--(2.5200,4.1829)--(2.6133,4.1784)--(2.7067,4.1355)--(2.8000,4.1667)--(2.8933,4.1493)--(2.9867,4.1457)--(3.0800,4.1658)--(3.1733,4.1508)--(3.2667,4.1658)--(3.3600,4.1553)--(3.4533,4.1817)--(3.5467,4.1541)--(3.6400,4.1589)--(3.7333,4.1949)--(3.8267,4.2033)--(3.9200,4.1700)--(4.0133,4.2174)--(4.1067,4.2024)--(4.2000,4.2114)--(4.2933,4.2081)--(4.3867,4.2270)--(4.4800,4.2231)--(4.5733,4.2231)--(4.6667,4.2723)--(4.7600,4.2864)--(4.8533,4.2522);
        \draw[blue](0.2800,7.2432)--(0.5600,6.0009)--(0.8400,5.1354)--(1.1200,4.8636)--(1.4000,4.6782)--(1.6800,4.5417)--(1.9600,4.4424)--(2.2400,4.3914)--(2.5200,4.3221)--(2.8000,4.2957)--(3.0800,4.2687)--(3.3600,4.2393)--(3.6400,4.2024)--(3.9200,4.2099)--(4.2000,4.1661)--(4.4800,4.1511)--(4.7600,4.1700)--(5.0400,4.1511)--(5.3200,4.1568)--(5.6000,4.1595)--(5.8800,4.1778)--(6.1600,4.1499)--(6.4400,4.1565)--(6.7200,4.1640);

        \draw[<->](7,8.5)--(0,8.5)--(0,12.1);
        \node[black] at (7.2,8.7) {\small{$step$}};
        \node[black] at (1,12.1) {\small{$loss \ of \ a$}};
        \node[red] at (3.5,9.1) {\small{$(19000,0.777)$}};
        \node[blue] at (6.5,9) {\small{$(57500,0.767)$}};
        \draw[red](0.0933,12.0147)--(0.1867,11.3628)--(0.2800,10.2404)--(0.3733,9.4914)--(0.4667,9.1395)--(0.5600,9.0141)--(0.6533,8.9447)--(0.7467,8.9038)--(0.8400,8.8787)--(0.9333,8.8451)--(1.0267,8.8211)--(1.1200,8.7911)--(1.2133,8.7854)--(1.3067,8.7857)--(1.4000,8.7655)--(1.4933,8.7638)--(1.5867,8.7644)--(1.6800,8.7418)--(1.7733,8.7291)--(1.8667,8.7340)--(1.9600,8.7194)--(2.0533,8.7126)--(2.1467,8.7109)--(2.2400,8.7071)--(2.3333,8.7167)--(2.4267,8.7179)--(2.5200,8.7068)--(2.6133,8.7009)--(2.7067,8.7118)--(2.8000,8.7086)--(2.8933,8.6998)--(2.9867,8.7011)--(3.0800,8.7027)--(3.1733,8.7131)--(3.2667,8.7066)--(3.3600,8.7088)--(3.4533,8.7047)--(3.5467,8.6993)--(3.6400,8.7047)--(3.7333,8.7175)--(3.8267,8.7138)--(3.9200,8.7141)--(4.0133,8.7164)--(4.1067,8.7182)--(4.2000,8.7155)--(4.2933,8.7205)--(4.3867,8.7446)--(4.4800,8.7234)--(4.5733,8.7260)--(4.6667,8.7431)--(4.7600,8.7386)--(4.8533,8.7522);
        \draw[blue] (0.2800,11.9251)--(0.5600,10.5975)--(0.8400,9.5296)--(1.1200,9.1381)--(1.4000,9.0052)--(1.6800,8.9134)--(1.9600,8.8667)--(2.2400,8.8372)--(2.5200,8.8086)--(2.8000,8.7931)--(3.0800,8.7617)--(3.3600,8.7591)--(3.6400,8.7424)--(3.9200,8.7559)--(4.2000,8.7141)--(4.4800,8.7122)--(4.7600,8.7175)--(5.0400,8.7153)--(5.3200,8.6989)--(5.6000,8.7183)--(5.8800,8.7089)--(6.1600,8.6929)--(6.4400,8.6954)--(6.7200,8.7013);
    \end{tikzpicture}
    \caption{The training curves of \textit{Task} $a$ and $main$ using \textcolor{red}{single model} and \textcolor{blue}{multitask model}. The optimum points are marked in the figures. Note that our batch size is 64.}
    \label{fg4}
\end{figure}
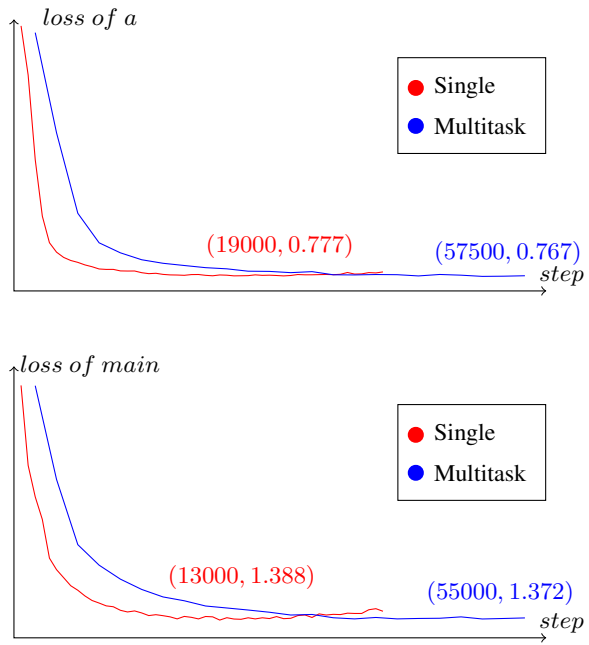

When training the multitask model, we jointly train five tasks in one model, so the efficiency of our strategy is an inevitable topic. Here in Figure \ref{fg4}, we demonstrate the training curves of \textit{Task} $a$ and $main$ using \textcolor{red}{single model} and \textcolor{blue}{multitask model}. 

We can clearly see that the convergence speed of multitask model is not five times slower than the single model. In fact, it only takes about three times of steps in \textit{Task} $a$ and four times in \textit{Task} $main$, for our multitask model to meet the optimum point compared with the single model. Also, the initial convergence speed in the first few steps of the single model is only about twice as fast as the joint model. Thus, in training we can say that the five tasks mutually benefit each other. In inference our multitask model takes about five times as long to generate a question.

\subsection{Different Reranking Losses}
\label{apbrl}
Besides the reranking losses defined in \ref{m32}, we also conduct another version which uses $\sum$ to aggregate single losses. We use Joint train+rerank+RS+SBS model to augment CoQA dataset and do DocNLI task, using $\prod$ and $\sum$ 036 respectively.
Table \ref{tbdrl} shows that the methods using $\prod$ gain almost the same performance as $\sum$.
\begin{table}[!h]
    \renewcommand\tabcolsep{6pt}
    \renewcommand\arraystretch{1.4}
    \centering
    \small
    \begin{tabular}{lccc}
    \toprule
    \textbf{CoQA} & \textbf{Bleu} & \textbf{Loss} & \textbf{F1$_{qa}$}\\
    \hline
    $\prod$  &32.01/47.86 &5.431/0.766 &72.49/81.98 \\
    $\sum$  &31.96/47.67 &5.404/0.756 &72.50/81.91 \\ 
    \bottomrule
    \textbf{DocNLI} & \textbf{Loss} & \textbf{F1$_{qa}$} & \textbf{F1$_{nli}$}\\
    \hline
    $\prod$  &2.503/3.457 &66.04/50.91 &50.01 \\
    $\sum$  &2.376/3.353 &66.19/51.19 &49.98 \\ 
    \bottomrule
    \end{tabular}
    \caption{Results of Joint train+rerank+RS+SBS model on augmenting CoQA dataset and DocNLI task, using different loss aggregation methods.}
    \label{tbdrl}
\end{table}

\subsection{Mathematically Analysis of Rationale Sampling}
\label{apbmrs}
Although the intuition of our rationale sampling is to use previous sentences and the Q-A pairs on $R_n$ to restore $R_n$, $\sum_{i=n'+1}^{n} (Q_i+A_i)$ is dependent on and logically connected with $\sum_{i=1}^{n'} (Q_i+A_i)$.
Also, since the information of $\sum_{i=1}^{n'} (Q_i+A_i)$ is totally contained in $\bigcup_{i=1}^{n'}R_i$, we might as well do the following transformation.
\begin{equation*}
\begin{split}
loss (R_n | \sum_{i=n'+1}^{n} (Q_i+A_i)+\bigcup_{i=1}^{n'}R_i, \theta) 
\\ \approx loss (R_n | \sum_{i=1}^{n} (Q_i+A_i)+\bigcup_{i=1}^{n'}R_i, \theta).
\end{split}
\end{equation*}

Also, since the information of $\sum_{i=n'+1}^{n}(Q_i+A_i)$ contribute not much to generate $\bigcup_{i=1}^{n'}R_i$, we can say that 
\begin{equation*}
\begin{split}
p(\bigcup_{i=1}^{n'}R_i|\sum_{i=1}^{n}(Q_i+A_i),\theta) 
\\ \approx p(\bigcup_{i=1}^{n'}R_i|\sum_{i=1}^{n'}(Q_i+A_i,\theta).
\end{split}
\end{equation*}
Then,

\begin{equation*}
\begin{split}
    & \quad loss (R_n | \sum_{i=n'+1}^{n} (Q_i+A_i)+\bigcup_{i=1}^{n'}R_i, \theta) 
    \\&\approx
    loss (R_n | \sum_{i=1}^{n} (Q_i+A_i)+\bigcup_{i=1}^{n'}R_i, \theta) 
    \\&= 
    -\frac{log\,p(R_n | \sum_{i=1}^{n} (Q_i+A_i)+\bigcup_{i=1}^{n'}R_i, \theta)}{m_n-m_{n'}}
    \\&=
    -\frac{1}{m_n-m_{n'}}[
    \\& \quad log\,p(R_n|\sum_{i=1}^{n} (Q_i+A_i)+\bigcup_{i=1}^{n'}R_i, \theta)
    \\&+ log\,p(\bigcup_{i=1}^{n'}R_i|\sum_{i=1}^{n}(Q_i+A_i),\theta)
    \\&- log\,p(\bigcup_{i=1}^{n'}R_i|\sum_{i=1}^{n}(Q_i+A_i),\theta)]
    \\&=
    -\frac{1}{m_n-m_{n'}} [log\,p(\bigcup_{i=1}^{n}R_i|\sum_{i=1}^{n}(Q_i+A_i),\theta) 
    \\&- log\,p(\bigcup_{i=1}^{n'}R_i|\sum_{i=1}^{n}(Q_i+A_i),\theta)]\ (use\ Eq.\ref{eq2})
    \\&\approx
    -\frac{1}{m_n-m_{n'}} [log\,p(\bigcup_{i=1}^{n}R_i|\sum_{i=1}^{n}(Q_i+A_i),\theta) 
    \\&- log\,p(\bigcup_{i=1}^{n'}R_i|\sum_{i=1}^{n'}(Q_i+A_i),\theta)]
    \\&= \frac{1}{m_n-m_{n'}} (m_nloss_{h_n} - m_{n'}loss_{h_{n'}})
    \triangleq a.
\end{split}
\end{equation*}

\subsection{Other Rationale Sampling Strategies}
\label{apbrs}
Besides the rationale sampling strategy in \ref{m33}, we also conduct two other versions. The first one is a constant function with a value of 0.3, as Eq.\ref{eqb1}. In the second version, we use the length of each rationale on behalf of its amount of information. We let $x$ mean the ratio between the current rationale length and the story length and make $kp$ linear related to $x$. Empirically, we set the slope to 3 and an upper bound of 0.75, as Eq.\ref{eqb2}. 

\begin{equation}
    kp=0.3.\label{eqb1}
\end{equation}

\begin{equation}
    kp=\left\{
    \begin{aligned}
    3x, \quad & \ 0\le x\le0.25 \\
    0.75, \quad & \ 0.25<x\le1 \\
    \end{aligned}
    \right.\label{eqb2}
\end{equation}

\begin{table}[!h]
    \renewcommand\tabcolsep{6pt}
    \renewcommand\arraystretch{1.4}
    \centering
    \small
    \begin{tabular}{lccc}
    \toprule
    \textbf{Tasks} & Eq.\ref{eqb1} & Eq.\ref{eqb2} & Ours \\
    \hline
    CoQA  &0.772/81.62 &0.762/81.88 &0.766/\textbf{81.98} \\
    SQuAD  &0.660/85.43 &0.651/\textbf{85.61} &0.691/85.59 \\ 
    DocNLI  & 2.382/49.12 &2.375/49.88 & 2.376/\textbf{49.98} \\
    \bottomrule
    \end{tabular}
    \caption{Results (F1$_{qa}$ for CoQA and SQuAD, F1$_{nli}$ for DocNLI) using different rationale sampling strategies.}
    \label{tb14}
\end{table}

Using these three rationale sampling methods, we train three sets of models on different tasks with full strategies. The results are in Table \ref{tb14}. We can see that the dynamic probability is more suitable than the constant value. Also, our strategy based on auxiliary \textit{Task} $h$ performs better than that based on sentence length. Specifically, it gets 0.1 points higher on CoQA and DocNLI and gets almost the same score on SQuAD.

\end{document}